# Detecting Visual Cues in the Intensive Care Unit and Association with Patient Clinical Status

Subhash Nerella[a,b], Ziyuan Guan[a], Andrea Davidson[a,c], Yuanfang Ren[a], Tezcan Baslanti[a], Brooke Armfield[a], Patrick Tighe[a,d], Azra Bihorac[a,c], Parisa Rashidi[a,b]

[a]Intelligent Critical Care Center, University of Florida, Gainesville, Florida, USA
[b]Biomedical Engineering, University of Florida, Gainesville, Florida, USA
[c]Division of Nephrology, Hypertension and Renal Transplantation, University of Florida, Gainesville, Florida, USA
[d]Department of Anesthesiology, University of Florida, Gainesville, Florida, USA

## Abstract

Intensive Care Units (ICU) provide close supervision and continuous care to patients with life-threatening conditions. However, continuous patient assessment in the ICU is still limited due to time constraints and the workload on healthcare providers. Existing patient assessments in the ICU such as pain or mobility assessment are mostly sporadic and administered manually, thus introducing the potential for human errors. Developing Artificial intelligence (AI) tools that can augment human assessments in the ICU can be beneficial for providing more objective and granular monitoring capabilities. For example, capturing the variations in a patient's facial cues related to pain or agitation can help in adjusting pain-related medications or detecting agitation-inducing conditions such as delirium. Additionally, subtle changes in visual cues during or prior to adverse clinical events could potentially aid in continuous patient monitoring when combined with high-resolution physiological signals and Electronic Health Record (EHR) data. In this paper, we examined the association between visual cues and patient condition including acuity status, acute brain dysfunction, and pain experienced. In this study we have processed 20,727,379 image frames obtained from 271 patients admitted to the ICUs at University of Florida Shands hospital. We leveraged our AU-ICU dataset with 107,064 frames collected in the ICU annotated with facial action units (AUs) labels by trained annotators. We developed a new "masked loss computation" technique that addresses the data imbalance problem by maximizing data resource utilization. We trained the model using our AU-ICU dataset in conjunction with three external datasets to detect 18 AUs. The SWIN Transformer model achieved 0.57 mean F1-score and 0.89 mean accuracy on the test set. Additionally, we performed AU inference on 634,054 frames to evaluate the association between facial AUs and clinically important patient conditions such as acuity status, acute brain dysfunction, and pain.

# Graphical Abstract

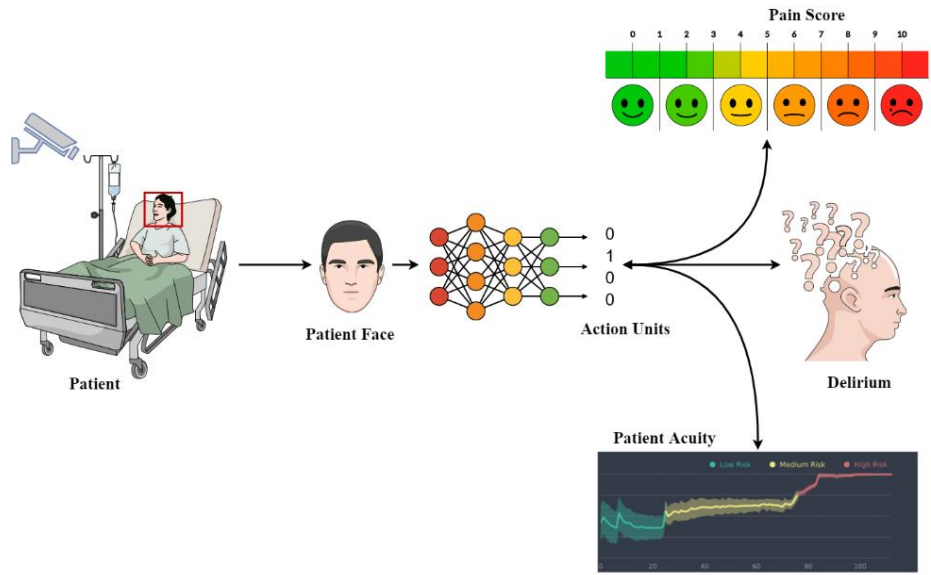

# Introduction

Every year, over 5.7 million critically ill patients are admitted to the Intensive Care Units (ICUs) in the United States. As a specialized unit, ICUs cater to patients with life-threatening illnesses by providing continuous care, supervision, life support equipment, and medication. Although ICUs are established for continuous monitoring, their functionality is still limited due to time and human resources constraints. ICU nurses spend significant time documenting or retrieving data from Electronic Health Record (EHR) systems, limiting their time left for direct patient interactions and observations [1-3]. Nonetheless, despite limited time, nurses still have to rely on many manual visual observations to evaluate different aspects of ICU patient care, such as pain in non-verbal patients, agitation in delirious patients, or mobility assessment. These repetitive visual examinations are burdensome and lead to burnout. It is estimated that a third of ICU nursing staff reported burnout [4]. Furthermore, visual information related to pain, brain dysfunction, and acuity evaluated can be only captured sporadically, every few hours at best [5, 6]. Nonetheless, it has been shown that the human face plays a vital role in nonverbal communication [7, 8], for example as a reflexive reaction to a painful experience [9], as an amimia expression in delirious patients, or as an indicator of deterioration in critically ill patients [10].

Pain Assessment. During ICU stay, 40-70% of patients experience moderate to severe pain [11-13]. Pain assessment is essential in the ICU to prevent deleterious complications, delirium onset, length of ICU stay, and mortality [14]. Currently, patient self-reported information is the gold standard for pain assessment. Self-reported pain scores are commonly captured through the Visual Analog Scale (VAS) [15, 16], or the Numeric Rating Scale (NRS) [17]. These scales are subjective to the individual and have a linear representation of pain; hence, they do not address the multidimensional aspects of pain. Furthermore, in many cases, critically ill patients cannot self-report pain for reasons such as being under mechanical ventilation, an altered mental state caused by the onset of delirium or dementia, and being under sedatives. In the case of non-verbal patients, ICU nurses resort to manual observation for pain assessment. Some of the nonverbal pain assessment tools employed by ICU nurses include the Non-Verbal Pain Scale (NVPS) [18], Behavioral Pain Scale (BPS) [19], and Critical Care Pain Observation tool (CPOT) [20]. Facial behavior analysis is a prominent factor in most pain assessment tools such as NVPS, BPS, and CPOT. However, the current facial behavior components of clinical assessments are only limited to a few facial cues (e.g., NVPS looks for grimace, tearing, frowning, and wrinkled forehead to assess pain) and lack granularity. Furthermore, these assessments must be manually administered, are prone to documentation errors, and have a significant lag between observation and documentation.

<u>Delirium</u>. Delirium is an acute neuropsychiatric clinical syndrome characterized by altered consciousness, cognition, inability to focus, disorientation, and sustained or shifted attention [21]. Delirium is linked to prolonged hospital stay, prolonged mechanical ventilation, increased mortality, and an annual cost of $164 billion in the United States [22, 23]. The incidence of delirium in the ICU can range from 45% to 87% [24]. The high prevalence rate of delirium in the ICU calls for timely detection of delirium. However, delirium assessment in the ICU requires manual administration, resulting in delayed detection. Many delirious patients appear frightened, confused or show no facial expression, called amimia [25]. Observing patient facial cues can be valuable for delirium assessment and management, by augmenting existing methods with further contextual information, as shown by Davoudi et al [26].

<u>Acuity</u>. Patient acuity level represents illness severity and the amount of nursing care the patient requires. Early detection of evolving illness severity can facilitate life-saving interventions and ensure optimum nursing resource utilization. Currently, the most common ICU assessment score to assess patient acuity is the Sequential Organ Failure Assessment (SOFA) score [27] or the National Early Warning Score (NEWS) [28]. Although these scores can provide reasonable patient acuity assessment, visual cues could potentially provide additional insights into changes in patient acuity. For example, Madrigal-Garcia and colleagues [10] showed that patterned facial expressions can be identified in deteriorating general ward patients.

Most prior research on facial behavior uses a facial anatomy-based action system referred to as the facial action coding system (FACS) [29]. The FACS system breaks down instant changes in facial expressions into individual facial action units (AUs). An AU is a contraction or relaxation of one or more facial muscles, which results in a visual appearance change on the face. The AUs must be manually coded by skilled professionals, which requires time-consuming and costly training methods and renders this strategy clinically unviable. An AI-driven AU detection system can overcome the limitations of the current manual observations to facilitate real-time assessments in the ICU.

In this paper, we developed an end-to-end AU detection pipeline to train a SWIN Transformer model on our AU-ICU dataset, which consists of an annotated collection of video frames collected from an uncontrolled ICU environment at the University of Florida (Figure 1). To address the data imbalance issue within our dataset, we incorporated external datasets during the training process. Subsequently, we performed inference using the trained SWIN [30] Transformer model on a substantial dataset of 634,508 patient frames, with the primary objective of detecting Action Units (AUs) in facial expressions. We computed the association between facial AUs detected on patient face images against ICU-assessed pain scores and patient ABD and acuity status phenotypes, computed as defined in [31, 32].

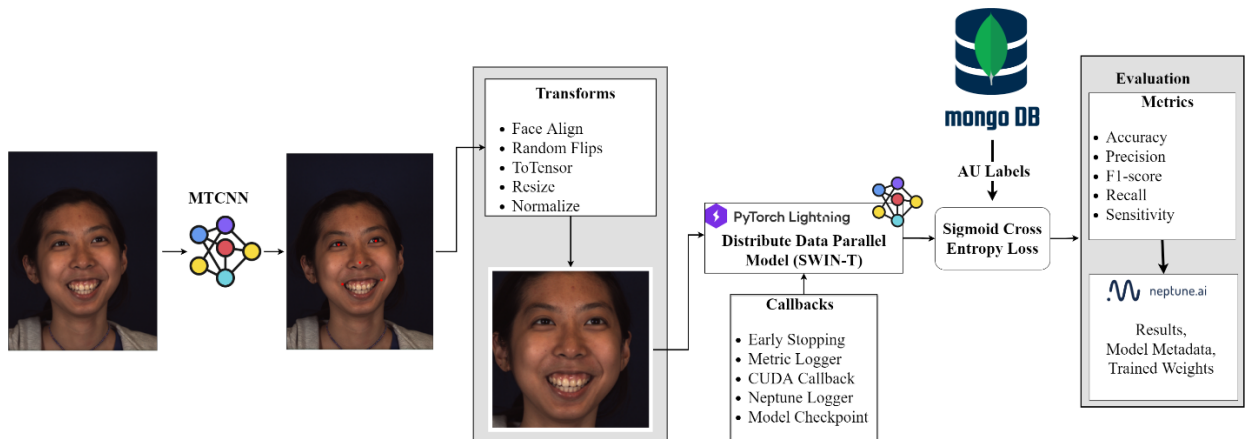

Figure 1. The end-to-end Facial Action Unit (AU) detection pipeline. The video frames of the patient's face are provided as input to the Multitask Cascaded Convolutional Network (MTCNN) network to detect facial landmarks. Facial landmarks are used to crop and align the face. Aligned faces are provided as input to train the SWIN transformer model. The model is trained on two GPUs using the distributed data-parallel training paradigm. Ground truth labels are obtained from our Mongo DB annotation dataset to compute the masked sigmoid cross entropy loss per AU. All the metrics, model metadata, and trained weights are logged to an MLOps tool (neptune.ai).

# Results

## Study Participants

All the data collected in this study is obtained from adult patients admitted to the surgical ICUs at the University of Florida Shands Health Hospital, Gainesville, Florida. Prior to study commencement, the study was reviewed and received approval from the University of Florida Institutional Review Board (UF-IRB) by IRB201900354 and IRB202101013. We obtained informed consent from all the study participants, and data collection strictly followed the UF-IRB guidelines and regulatory standards. We used the exclusion criteria and development and validation cohorts presented in the cohort diagram (Figure 2) to train the SWIN Transformer model. In Table 1, we illustrate the cohort characteristics of patients whose data is used in this study.

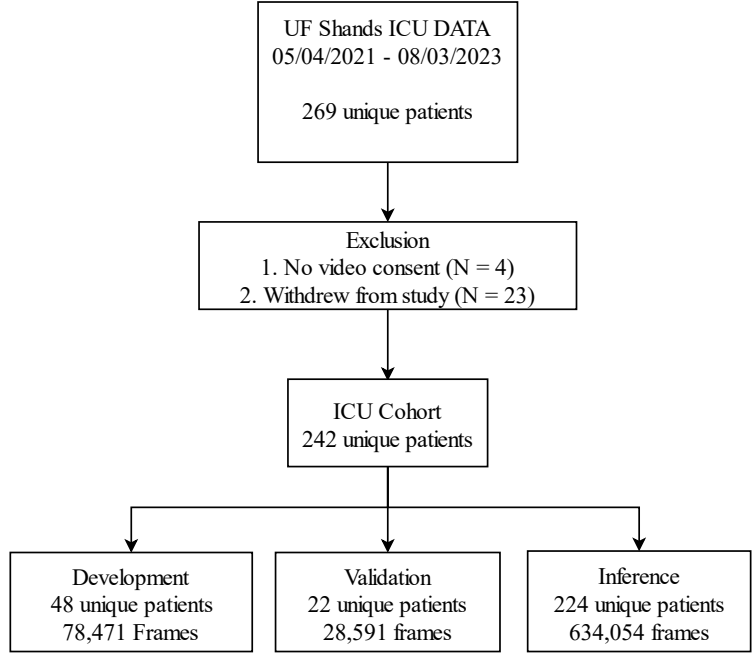

Figure 2. Overview of cohort selection criteria and derivation of development, validation, and inference cohort.

Table 1. Patient Characteristics Table

| Participants (n = 271, number of frames = 20,727,379) | |
|---|---|
| Age, median (IQR) | 71 (47,62) |
| Gender, number (%) | Male 143 (59) <br> Female 100 (41) |
| Race, number (%) | White, 197 (80) <br> African American, 25 (8) <br> Asian or Pacific Islander, 6 <br> Other, 6 (12) |
| ICU LOS, median (IQR) | 9.7 (4.8,24.1) |
| Hospital LOS, median (IQR) | 19.8 (9.5,36.2) |
| Brain dysfunction | Normal (147) <br> Coma/Delirium (47) |
| Patient Acuity Status | Stable (117) <br> Unstable (97) |
| Death, number(%) | |

LOS: Length Of Stay, IQR: Inter Quartile Range (.25,.75)

## Masked Loss Computation

AU detection is a challenging problem due to the variability of facial expressions in diverse cultures and individuals. Facial occlusions and changing lighting conditions can affect the AU detection accuracy. AU detection on patients in the ICU is further challenging. Moreover, AU-labeled datasets are unbalanced with a disproportionate presence of some AUs over others in the dataset. Our AU-ICU dataset is also highly imbalanced. To address this data imbalance and maximum available data resource utilization we propose masked loss computation.

|        | AU 1 | AU 4 | AU 6 | AU 9 | AU 14 | AU 20 | AU 27 | AU 43 |
|--------|------|------|------|------|-------|-------|-------|-------|
| ICU-AU | -1   | 1    | 0    | 0    | -1    | 0     | 1     | 0     |
| BP4D   | 0    | 0    | 1    | -1   | 0     | -1    | -1    | -1    |
| DISFA  | 1    | 0    | 1    | 0    | -1    | 1     | -1    | -1    |
| UNBC   | -1   | 0    | 0    | -1   | -1    | 0     | -1    | 1     |

(A)

Mask

| 0 | 1 | 1 | 1 | 0 | 1 | 1 | 1 |
| 1 | 1 | 1 | 0 | 1 | 0 | 0 | 0 |
| 1 | 1 | 1 | 1 | 0 | 1 | 0 | 0 |
| 0 | 1 | 1 | 0 | 0 | 1 | 0 | 1 |

(B)

Figure 3. Masked loss computation. (A) AUs absent in the dataset are given -1 as a dummy ground truth label. Color coding green represents AUs present and red represents AUs absent in the datasets. (B) Mask array multiplied with loss array to prevent faulty loss computed with dummy label from backpropagating gradients into the network.

$$L = y \cdot \log(\hat{y}) + (1 - y) \cdot \log(1 - \hat{y}) \qquad (1)$$

Masked loss computation can be used to simultaneously train multiple datasets, even when they lack common class labels. The ground truth labels in these datasets use value 1 to indicate AU is present and 0 to represent its absence in the image. Additionally, we introduced a -1 dummy ground truth to represent that the AU is not labeled for that specific dataset as shown in Figure 3. We use binary cross entropy (BCE) loss per image per AU. The BCE loss, as depicted in equation 1 is applied in the case of binary labels (0 or 1). The BCE loss is a summation of two terms, one of which always equals 0 depending on the ground truth label. However, the introduction of the dummy label -1 for AUs absent in the dataset results in a faulty loss computation. Backpropagating gradient based on this faulty loss will adversely affect the performance of the network. To prevent backpropagating this erroneous loss computed from these dummy labels we introduce the mask shown in Figure 3 that assumes value 1 if the AU is labeled in the data and 0 otherwise. We perform an element-wise multiplication between the loss that is computed per image per AU with the mask and then perform reduction with the average operation. This approach ensures the erroneous loss is computed with the ground truth label as -1 as in not backpropagated into the network.

## Training

We used four different datasets BP4D, DISFA, UNBC, and our in-house AU-ICU dataset to train the SWIN Transformer model. The combined dataset containing 432,919 face images obtained from 163 participants was split into a training set with 317,340 images from 116 participants and a test set with 115,579 frames from 47 participants. The training set comprises 78,471 images obtained from 48 patients and 28,591 images obtained from 22 patients. No participant data is common between training and test sets. We used the SWIN transformer base variant with 86.8 million parameters obtained from the Hugging Face library for training. The model is trained for 10 epochs along with earlystopping callback. We used the Adam [33] optimizer and two Nvidia 2080 TI GPUs to train the model. The model is trained with binary cross-entropy loss using the masked loss computation technique. We use an end-to-end AU detection pipeline to train the model. In Table 2 we report the performance of the SWIN transformer model on the test partition.

Table 2. Trained SWIN transformer performance on the test dataset.

| AU | Description | F1-score | Accuracy |
| --- | --- | --- | --- |
| AU1 | Inner Brow Raiser | 0.54 | 0.86 |
| AU2 | Outer Brow Raiser | 0.56 | 0.88 |
| AU4 | Brow Lowerer | 0.42 | 0.91 |
| AU6 | Cheek Raiser | 0.78 | 0.9 |
| AU7 | Lid Tightener | 0.76 | 0.88 |
| AU9 | Nose wrinkler | 0.36 | 0.97 |
| AU10 | Upper Lip Raiser | 0.84 | 0.9 |
| AU12 | Lip Corner Puller | 0.85 | 0.92 |
| AU14 | Dimpler | 0.58 | 0.63 |
| AU15 | Lip Corner Depressor | 0.45 | 0.93 |
| AU17 | Chin Raiser | 0.62 | 0.85 |
| AU20 | Lip Stretcher | 0.1 | 0.98 |
| AU23 | Lip Tightener | 0.5 | 0.83 |
| AU24 | Lip Pressor | 0.43 | 0.89 |
| AU25 | Lips part | 0.83 | 0.9 |
| AU26 | Jaw Drop | 0.55 | 0.91 |
| AU27 | Mouth Stretch | 0.42 | 0.99 |
| AU43 | Eyes Closed | 0.68 | 0.8 |
|  | Mean | 0.57 | 0.89 |

## Clinical Association

Our objective for this study is to investigate the association between patient facial cues and their clinical status. We collected the pain score, acuity status, and brain dysfunction status of the patient during their ICU stay. The trained SWIN Transformer model was used to run inference on 634,504 frames obtained from 242 patients to obtain facial AUs. We selectively included the AUs on which the model achieved at least 0.5 F1-score and 0.8 accuracy value on the test. The AU associations for a target class are computed as a ratio of frames where the AU is detected to the total number of frames considered. We also ran mixed effects logistic regression with AUs as independent variables and patients as random effects to evaluate the statistical significance of AUs against the clinical variables.

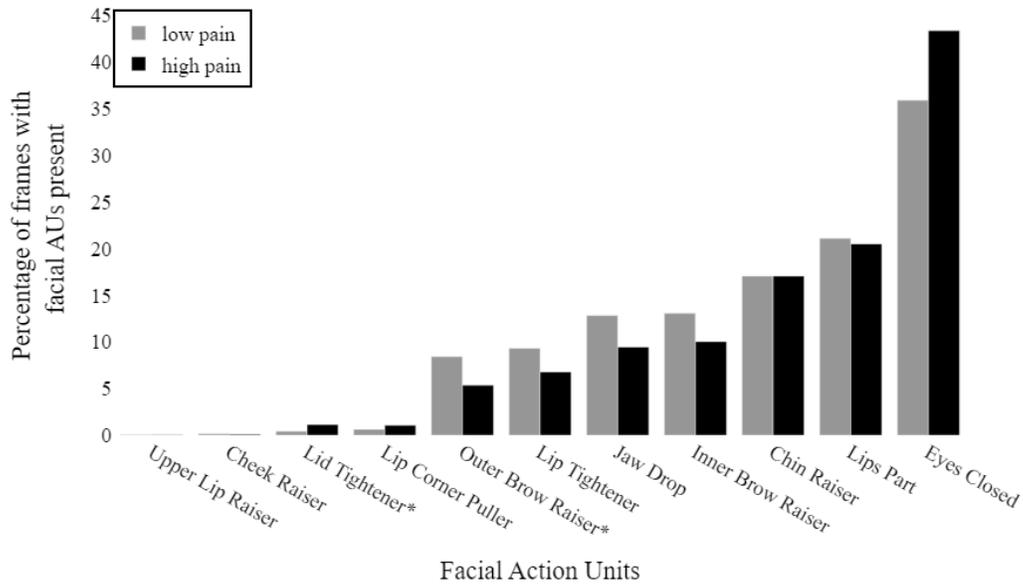

Figure 4. AU presence vs patient-reported pain score level. This bar chart shows the percentage of frames with the AU present to the total number of frames collected during patient-reported low pain (pain score <4) and high pain (pain score>=4) periods. *shows statistically significant AU presence difference between low pain and high pain periods.

The patient-reported pain score is categorized into two classes low pain (pain score <4) and high pain (pain score >=4). The criteria for this distinction was based on the DVPRS pain scale rating which categorized pain scores greater than or equal to 4 as interfering with usual activities. The facial AU presence detected in the video frames is compared against the patient-reported pain score categories shown in Figure 4. The mixed effect modeling showed AU2 (outer brow raiser) showed a statistically significant negative association ($P < 0.05$) with a pain score greater than 4, while AU 7 (Lid Tightener) showed a significant positive association ($P \sim 0.05$) with pain score greater than equal to 4. Although we observed a positive association with high pain scores from other AUs, these associations did not reach statistical significance.

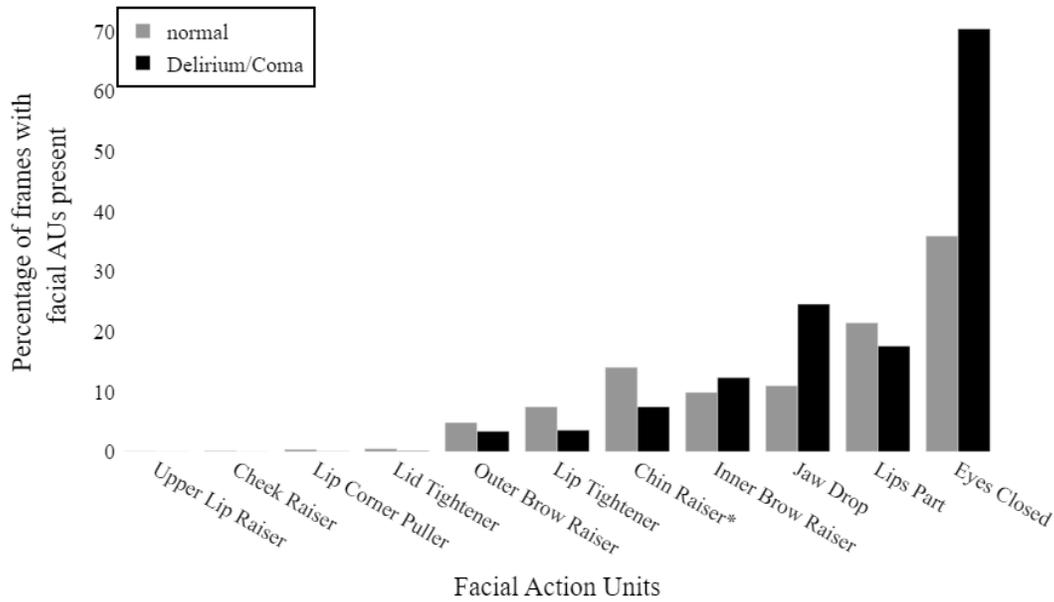

Figure 5. AU presence vs acute brain dysfunction status of the patient. This bar chart shows the percentage of frames with the AU present to the total number of frames collected during patient brain dysfunction status assessed by clinical staff as normal and abnormal(coma/delirium). *shows statistically significant AU presence difference between normal and abnormal patient brain dysfunction status.

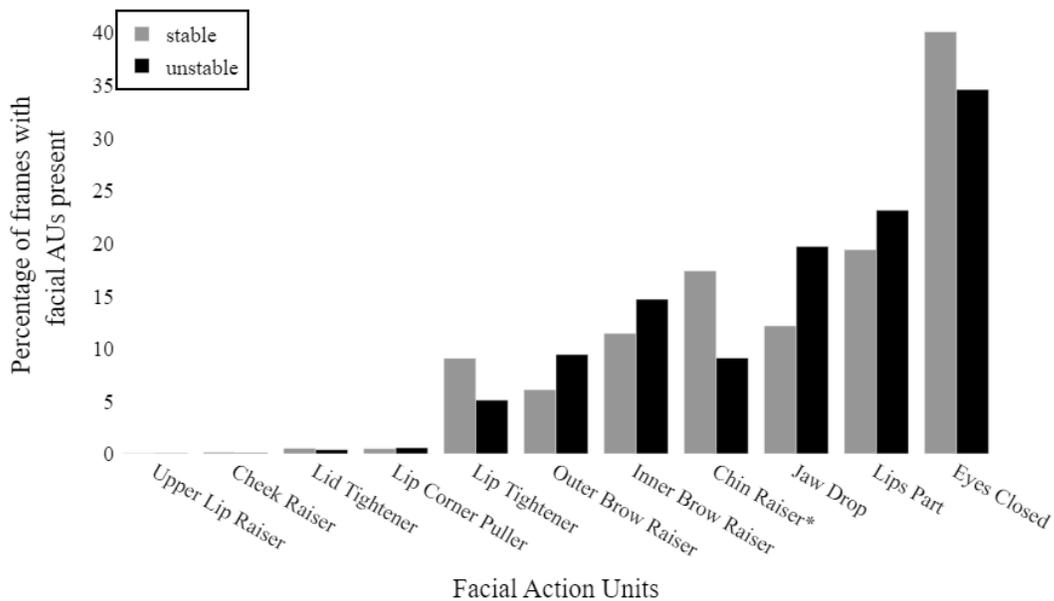

Figure 6. AU presence vs acuity status of the patient. This bar chart shows the percentage of frames with the AU present to the total number of frames collected during patient acuity status assessed by clinical staff as stable or unstable. *shows statistically significant AU presence difference between stable and unstable patient acuity status.

The patient's brain dysfunction status was categorized into normal, delirium, or coma based on the clinical assessments. Patient brain status is categorized into normal and abnormal (delirium/coma) and compared against the facial AU presence shown in Figure 5. Mixed effects logistic regression modeling revealed a significant (p<0.05) negative association between AU 17 (chin raiser) with abnormal brain status.

The comparison between facial AU presence and patient acuity status is shown in Figure 6. Mixed-effects logistic regression modeling on stable and unstable patient acuity levels against AU presence showed that AU 2 (outer brow raiser) has a positive association (p = 0.06) with unstable acuity state, while AU 17 (chin raiser) (p< 0.05), and AU23 (Lip Tightener) (p=0.07) showed a negative association with unstable acuity state.

# Discussion

In this study, we studied the feasibility of autonomous patient face assessment and evaluated the association between patient clinical status and pain experienced against the predicted facial cues. This work is first of its kind to collect facial videos in the ICU and label them for facial AUs. We used a SWIN transformer that uses a shifted window attention mechanism for the facial AU detection objective. To overcome the limitations of AU imbalance we trained our dataset together with external datasets BP4D, DISFA, and UNBC. While all four datasets employed in this study include annotations for facial AUs, there is limited overlap in the specific AUs they encompass. We introduced the masked loss computation technique to simultaneously train the model on multiple datasets. By training on multiple datasets using the masked loss computation, we were able to augment the quantity of data samples available for less represented Action Units (AUs), thereby effectively mitigating the data imbalance issue.

In order to find the association between patient clinical status and patient facial cues it is important to identify the AUs that are specific or sensitive to the pain experience, brain status, and acuity status. We used the mixed effects logistic regression to discern the statistically significant differences in AU presence between low pain vs high pain, acute brain dysfunction status normal vs (delirium/coma), and acuity status stable vs unstable videos. By observing the coefficients of mixed effects logistic regression, positive and negative presence association of AUs with clinical status can be determined. This statistical analysis revealed the AU7 (Lid Tightener) showed a statistically significant positive association with the high pain category, a noteworthy finding given that AU7 is one of the components of computing pain score in the Prkachin Solomon Pain Index (PSPI) [34] score. The PSPI score represents pain as a combination of AU-4 (Brow Lowerer), AU-6 (Cheek Raiser), AU-7 (Lid Tightener), AU-9 (Nose Wrinkler), AU-10 (Upper Lip Raiser), and AU-43 (Eyes Closed).

Our study has a few limitations. Although the bar charts (Figures 4-6) showed AU presence difference between the two groups, statistical analysis did not show significant differences. To run the statistical analysis, we consider each video as a data point. We observed the face detection module in the data processing pipeline could not accurately detect patient faces in the videos especially when there is life support medical equipment (ventilator, feeding tube) present on the patient's face. ICU patients under critical conditions are more likely to have medical equipment. This limitation resulted in few to no frames extracted from videos collected during the adverse clinical status of the patient. To mitigate this issue, we will train our own YOLO model to perform patient face and key face landmark detection by training the model on labeled data collected in the ICU. We developed a data annotation to generate labeled data to train the face detection model in the ICU. Another limitation, our data collection period is limited to two days and does not cover the entire duration of the patient in the ICU. Consequently, our dataset lacks records for numerous instances in which patients experienced adverse clinical events.

Privacy is a major concern when patient faces are recorded using video cameras. To ensure data security all the data used/collected in our research is securely stored, processed, and analyzed on a dedicated server isolated from the internet, accessible only through the University of Florida health VPN. We have rigorously adhered to all Institutional Review Board (IRB) guidelines, as well as state and federal regulations, to uphold the privacy and confidentiality of patient data included in our study. Moreover, the data has been de-identified to ensure that researchers have no access to any identifying patient information. In the future, the model will be for real-time AU detection to continuously infer patient facial cues without storing any patient face videos. This real-time inference, derived from our facial AU detection model, will subsequently be integrated into our patient risk prediction model.

# Methods

## Data Acquisition

We used an Amcrest camera (model IP2M-841W-V3) full-HD with a $90^0$ field of view to collect videos at 30 frames per second (fps) in the ICU. The camera is mounted on a standalone cart that can be easily maneuvered in and out of the ICU without causing any disruptions to the regular patient care activities within the unit. The camera was used solely for video collection, with no audio recording capability. The camera is zoomed onto the patient's bed to ensure we only capture patient faces from whom we have obtained prior consent. Additionally, the cart was outfitted with a computer and a monitor, and a user-friendly graphical interface was developed specifically for ICU staff and clinical coordinators to control the camera. This interface allows them to start, pause, and terminate the recording process at any time, either at the patient's request or during medical procedures, as required. The data collected was locally stored on the cart, and subsequently transferred to the central server. To maintain data security and organization, the videos were encrypted and systematically organized into patient-specific folders on the server.

## Data Processing

We implemented a data processing pipeline depicted in Figure 7 for extracting patient faces from the videos captured in the ICU. We selectively chose the videos within an hour time window of the patient-reported pain timestamp. Using the FFmpeg [35] multimedia processing tool, individual image frames are extracted from the videos at the rate of 1 frame per second. Subsequently, we used the Multitask Cascaded

Convolutional Network (MTCNN) [36] to detect the faces and crop the faces in the extracted frames. The cropped face images are assigned to trained annotators for labeling facial AUs.

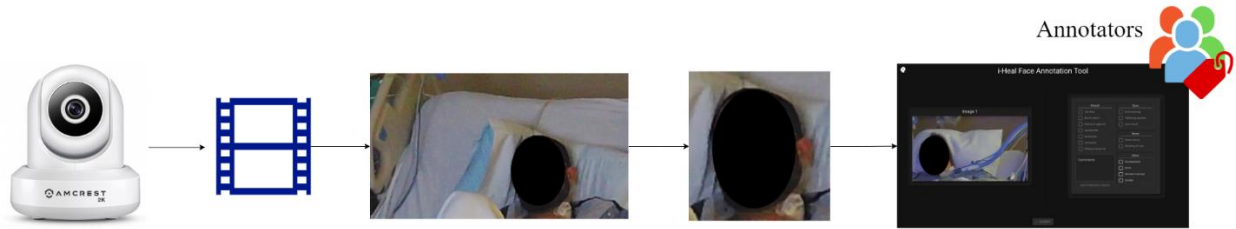

Figure 7. Patient facial video processing pipeline. The videos are processed to extract individual frames at 1fps. Patient faces are detected and cropped from the frames and subsequently assigned to trained annotators to label facial AUs.

Table 3. Facial AUs present in BP4D, DISFAPlus, and AU-ICU Datasets

| AU | Description | BP4D [37] | DISFA [38] | UNBC [39] | AU-ICU (ours) |
|---|---|---|---|---|---|
| 1 | Inner Brow Raiser | ✓ | ✓ | | |
| 2 | Outer Brow Raiser | ✓ | ✓ | | |
| 4 | Brow Lowerer | ✓ | ✓ | ✓ | ✓ |
| 5 | Upper Lid Raiser | | ✓ | | |
| 6 | Cheek Raiser | ✓ | ✓ | ✓ | ✓ |
| 7 | Lid Tightener | ✓ | | ✓ | ✓ |
| 9 | Nose wrinkler | | ✓ | ✓ | ✓ |
| 10 | Upper Lip Raiser | ✓ | | ✓ | ✓ |
| 12 | Lip Corner Puller | ✓ | ✓ | ✓ | ✓ |
| 14 | Dimpler | ✓ | | | |
| 15 | Lip Corner Depressor | ✓ | ✓ | ✓ | |
| 17 | Chin Raiser | ✓ | ✓ | | |
| 20 | Lip Stretcher | | ✓ | ✓ | ✓ |
| 23 | Lip Tightener | ✓ | | | |
| 24 | Lip Pressor | ✓ | | | ✓ |
| 25 | Lips part | | ✓ | ✓ | ✓ |
| 26 | Jaw Drop | | ✓ | ✓ | ✓ |
| 27 | Mouth Stretch | | | ✓ | ✓ |
| 43 | Eyes Closed | | | ✓ | ✓ |

## Data Annotation

We developed a proprietary, secure web-based annotation tool tailored towards annotating facial AUs on patient face images. Four annotators were carefully selected and trained on the FACS system prior to annotating patient images. To ensure scientific rigor, annotators were evaluated on sample face images before annotation. Each annotator carried out individual annotations using the dedicated annotation tool, which is integrated with a MongoDB database. We also hired an annotation manager to regularly assess the quality of annotation and efficiency of annotators. The annotation manager reviews the labels by each annotator and provides feedback to enhance their performance. We have a total of 107,064 frames annotated obtained from 70 patients. Table 3 shows AUs labeled in our dataset and other external datasets.

## External Datasets

**BP4D** [37] dataset consists of approximately 140,000 face image frames obtained from a diverse group of 41 young adult participants (23 female, 18 male). Eight different tasks were conceived to elicit eight different emotions in participants. The face image frames are AU labeled using the FACS system and 49 facial landmarks.

**DISFA** [38] dataset comprises frames obtained from 15 male and 12 female participants. The videos are annotated for 12 facial AUS and face landmarks. Notably, these annotations include the intensity of each AU, assessed on a scale ranging from 0 (indicating the absence of the AU) to 5 (representing the highest intensity of the AU). The dataset encompasses 12 AUs, as detailed in Table 2. In line with the methodology of prior studies [13, 14], AUs are considered present if the annotated AU intensity is equal to or greater than equal to 2.

**UNBC** [39] dataset is a collection of 200 video sequences obtained from 129 adult participants (63 male, 66 female) with shoulder pain. A subset of 48,398 frames obtained from 25 participants were coded with FACS facial AUs.

## SWIN Transformer model

We used the SWIN transformer shown in Figure 8 to train our AU-ICU dataset along with BP4D, DISFA, and UNBC datasets. SWIN model uses a transformer backbone and used a unique shifted window attention mechanism that reduces the attention computational complexity from quadratic to linear with respect to image size. In our previous work [40], we compared the performance of the SWIN Transformer against the Vision transformer (ViT) [41] and JAA-Net model [42]. In summary, the SWIN Transformer effectively struck a balance between model performance and training speed on AU detection problem.

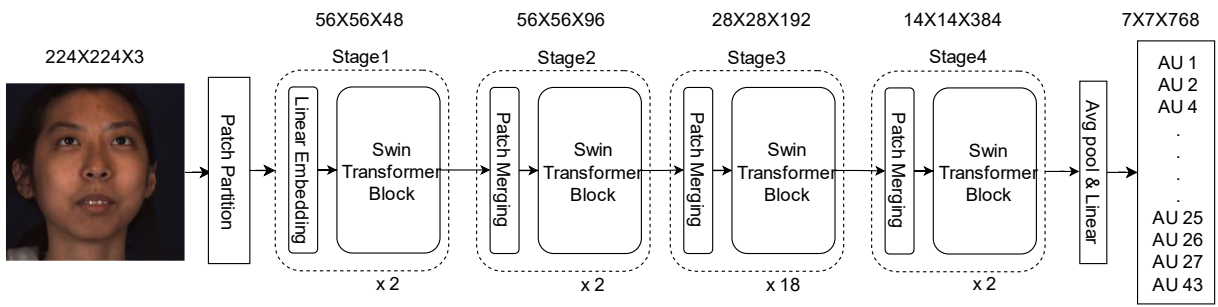

Figure 8. SWIN Transformer base variant architecture.

## Clinical Assessments

**Pain assessment**. We obtained the pain scores recorded by ICU nurses from the UF Integrated data repository. The pain scores are evaluated on the Defense and Veterans Pain Rating Scale (DVPRS) which ranges from 0-10. The DVPRS scale is basic traffic light color coded where green represents mild pain (0-4); yellow represents moderate pain (5-6), and red for severe pain (7-10).

## Computable Phenotypes

**Patient acuity** status is a representation of the amount of nursing resources required by a patient in the ICU. In this study, we generated acuity labels for every four-hour intervals starting from patient admission into the ICU based on the criteria proposed by Ren et al. [31] for developing computable phenotypes for acuity status. Every interval patient acuity status is categorized into stable or unstable categories. Patients requiring at least one of the life support therapies that include continuous renal replacement therapy, mechanical ventilation, vasopressors, or blood transfusion are considered unstable or otherwise stable.

**Acute Brain Dysfunction** (ABD) is a critical brain dysfunction that often manifests as delirium or coma among patients admitted to ICU. Patient ABD status is classified into comatose (coma), delirium, and normal states every 12 hours using the criteria proposed by Ren et al. [32] to characterize patient brain dysfunction using computable phenotypes. ABD status is computed based on the Confusion Assessment Method (CAM) [43], Richmond Agitation Sedation Scale (RASS) [44], and Glasgow Coma Scale (GCS) [45] assessment scores of patients administered by ICU nurses.

# Acknowledgement

A.B, P.R., and T.B. were supported by NIH/NINDS R01 NS120924, NIH/NIBIB R01 EB029699. P.R. was also supported by NSF CAREER 1750192.

# References

[1] D. H. Wong, Y. Gallegos, M. B. Weinger, S. Clack, J. Slagle, and C. T. Anderson, "Changes in intensive care unit nurse task activity after installation of a third-generation intensive care unit information system," *Critical care medicine,* vol. 31, no. 10, pp. 2488-2494, 2003.

[2] L. Mamykina, D. K. Vawdrey, and G. Hripcsak, "How do residents spend their shift time? A time and motion study with a particular focus on the use of computers," *Academic medicine: journal of the Association of American Medical Colleges,* vol. 91, no. 6, p. 827, 2016.

[3] L. W. Higgins *et al.*, "Hospital nurses' work activity in a technology-rich environment," *Journal of nursing care quality,* vol. 32, no. 3, pp. 208-217, 2017.

[4] M. Verdon, P. Merlani, T. Perneger, and B. Ricou, "Burnout in a surgical ICU team," *Intensive care medicine,* vol. 34, no. 1, pp. 152-156, 2008.

[5] S. M. Parry *et al.*, "Assessment of impairment and activity limitations in the critically ill: a systematic review of measurement instruments and their clinimetric properties," *Intensive care medicine,* vol. 41, no. 5, pp. 744-762, 2015.

[6] A. Thrush, M. Rozek, and J. L. Dekerlegand, "The clinical utility of the functional status score for the intensive care unit (FSS-ICU) at a long-term acute care hospital: a prospective cohort study," *Physical therapy,* vol. 92, no. 12, pp. 1536-1545, 2012.

[7]     F. De la Torre and J. Cohn, "Guide to Visual Analysis of Humans: Looking at People, chapter Facial Expression Analysis," *Springer,* vol. 9, pp. 31-33, 2011.
[8]     P. Ekman, W. V. Friesen, M. O'Sullivan, and K. Scherer, "Relative importance of face, body, and speech in judgments of personality and affect," *Journal of personality and social psychology,* vol. 38, no. 2, p. 270, 1980.
[9]     K. D. Craig, K. M. Prkachin, and R. E. Grunau, "The facial expression of pain," 2011.
[10]    M. I. Madrigal-Garcia, M. Rodrigues, A. Shenfield, M. Singer, and J. Moreno-Cuesta, "What faces reveal: A novel method to identify patients at risk of deterioration using facial expressions," *Critical Care Medicine,* vol. 46, no. 7, pp. 1057-1062, 2018.
[11]    N. A. Desbiens and A. W. Wu, "Pain and suffering in seriously ill hospitalized patients," *Journal of the American Geriatrics Society,* vol. 48, no. S1, pp. S183-S186, 2000.
[12]    D. T. Li and K. Puntillo, "A pilot study on coexisting symptoms in intensive care patients," *Applied Nursing Research,* vol. 19, no. 4, pp. 216-219, 2006.
[13]    G. Chanques *et al.*, "Psychometric comparison of three behavioural scales for the assessment of pain in critically ill patients unable to self-report," *Critical Care,* vol. 18, pp. 1-12, 2014.
[14]    F. Tennant, "Complications of uncontrolled, persistent pain," *Pract Pain Manag,* vol. 4, no. 1, pp. 11-14, 2004.
[15]    P. E. Bijur, W. Silver, and E. J. Gallagher, "Reliability of the visual analog scale for measurement of acute pain," *Academic emergency medicine,* vol. 8, no. 12, pp. 1153-1157, 2001.
[16]    H. Breivika, "Fifty years on the Visual Analogue Scale (VAS) for pain-intensity is still good for acute pain. But multidimensional assessment is needed for chronic pain," *Scandinavian journal of pain,* vol. 11, no. 1, pp. 150-152, 2016.
[17]    C. L. Von Baeyer, L. J. Spagrud, J. C. McCormick, E. Choo, K. Neville, and M. A. Connelly, "Three new datasets supporting use of the Numerical Rating Scale (NRS-11) for children's self-reports of pain intensity," *PAIN®,* vol. 143, no. 3, pp. 223-227, 2009.
[18]    A. M. Kabes, J. K. Graves, and J. Norris, "Further validation of the nonverbal pain scale in intensive care patients," *Critical care nurse,* vol. 29, no. 1, pp. 59-66, 2009.
[19]    J.-F. Payen *et al.*, "Assessing pain in critically ill sedated patients by using a behavioral pain scale," *Critical care medicine,* vol. 29, no. 12, pp. 2258-2263, 2001.
[20]    C. Gélinas, L. Fillion, K. A. Puntillo, C. Viens, and M. Fortier, "Validation of the critical-care pain observation tool in adult patients," *American Journal of Critical Care,* vol. 15, no. 4, pp. 420-427, 2006.
[21]    T. G. Fong, S. R. Tulebaev, and S. K. Inouye, "Delirium in elderly adults: diagnosis, prevention and treatment," *Nature Reviews Neurology,* vol. 5, no. 4, pp. 210-220, 2009.
[22]    J. I. Salluh *et al.*, "Outcome of delirium in critically ill patients: systematic review and meta-analysis," *bmj,* vol. 350, 2015.
[23]    D. L. Leslie, E. R. Marcantonio, Y. Zhang, L. Leo-Summers, and S. K. Inouye, "One-year health care costs associated with delirium in the elderly population," *Arch Intern Med,* vol. 168, no. 1, pp. 27-32, Jan 14 2008, doi: 10.1001/archinternmed.2007.4.
[24]    R. Cavallazzi, M. Saad, and P. E. Marik, "Delirium in the ICU: an overview," *Annals of intensive care,* vol. 2, no. 1, pp. 1-11, 2012.
[25]    A. Granberg-Axèll, I. Bergbom, and D. Lundberg, "Clinical signs of ICU syndrome/delirium: an observational study," *Intensive and Critical Care Nursing,* vol. 17, no. 2, pp. 72-93, 2001.
[26]    A. Davoudi *et al.*, "The Intelligent ICU Pilot Study: Using Artificial Intelligence Technology for Autonomous Patient Monitoring," *ArXiv preprint arXiv:1804.10201,*
[27]    J.-L. Vincent *et al.*, "The SOFA (Sepsis-related Organ Failure Assessment) score to describe organ dysfunction/failure: On behalf of the Working Group on Sepsis-Related Problems of the European Society of Intensive Care Medicine (see contributors to the project in the appendix)," ed: Springer-Verlag, 1996.
[28]    L. RCoP, "National Early Warning Score (NEWS): standardising the assessment of acute-illness severity in the NHS," *Report of working party. London: Royal College of Physicians,* 2012.
[29]    P. Ekman, W. V. Friesen, and J. C. Hager, "Facial action coding system: The manual on CD ROM," *A Human Face, Salt Lake City,* pp. 77-254, 2002.
[30]    Z. Liu *et al.*, "Swin transformer: Hierarchical vision transformer using shifted windows," *arXiv preprint arXiv:2103.14030,* 2021.

[31] Y. Ren *et al.*, "Development of Computable Phenotype to Identify and Characterize Transitions in Acuity Status in Intensive Care Unit," *arXiv preprint arXiv:2005.05163,* 2020.
[32] Y. Ren *et al.*, "Computable Phenotypes to Characterize Changing Patient Brain Dysfunction in the Intensive Care Unit," *arXiv preprint arXiv:2303.05504,* 2023.
[33] D. P. Kingma and J. Ba, "Adam: A method for stochastic optimization," *arXiv preprint arXiv:1412.6980,* 2014.
[34] K. M. Prkachin and P. E. Solomon, "The structure, reliability and validity of pain expression: Evidence from patients with shoulder pain," *Pain,* vol. 139, no. 2, pp. 267-274, 2008.
[35] "FFmpeg Developers. (2016). ffmpeg tool (Version be1d324) [Software].", ed.
[36] K. Zhang, Z. Zhang, Z. Li, and Y. Qiao, "Joint face detection and alignment using multitask cascaded convolutional networks," *IEEE signal processing letters,* vol. 23, no. 10, pp. 1499-1503, 2016.
[37] X. Zhang *et al.*, "Bp4d-spontaneous: a high-resolution spontaneous 3d dynamic facial expression database," *Image and Vision Computing,* vol. 32, no. 10, pp. 692-706, 2014.
[38] S. M. Mavadati, M. H. Mahoor, K. Bartlett, P. Trinh, and J. F. Cohn, "Disfa: A spontaneous facial action intensity database," *IEEE Transactions on Affective Computing,* vol. 4, no. 2, pp. 151-160, 2013.
[39] P. Lucey, J. F. Cohn, K. M. Prkachin, P. E. Solomon, and I. Matthews, "Painful data: The UNBC-McMaster shoulder pain expression archive database," in *Face and Gesture 2011*, 2011: IEEE, pp. 57-64.
[40] S. Nerella, K. Khezeli, A. Davidson, P. Tighe, A. Bihorac, and P. Rashidi, "End-to-End Machine Learning Framework for Facial AU Detection in Intensive Care Units," *arXiv preprint arXiv:2211.06570,* 2022.
[41] A. Dosovitskiy *et al.*, "An image is worth 16x16 words: Transformers for image recognition at scale," *arXiv preprint arXiv:2010.11929,* 2020.
[42] Z. Shao, Z. Liu, J. Cai, and L. Ma, "JAA-Net: joint facial action unit detection and face alignment via adaptive attention," *International Journal of Computer Vision,* vol. 129, no. 2, pp. 321-340, 2021.
[43] E. W. Ely *et al.*, "Evaluation of delirium in critically ill patients: validation of the Confusion Assessment Method for the Intensive Care Unit (CAM-ICU)," *Critical care medicine,* vol. 29, no. 7, pp. 1370-1379, 2001.
[44] C. N. Sessler *et al.*, "The Richmond Agitation–Sedation Scale: validity and reliability in adult intensive care unit patients," *American journal of respiratory and critical care medicine,* vol. 166, no. 10, pp. 1338-1344, 2002.
[45] G. Teasdale and B. Jennett, "Assessment of coma and impaired consciousness: a practical scale," *The Lancet,* vol. 304, no. 7872, pp. 81-84, 1974.